\documentclass[10pt,twocolumn,letterpaper]{article}
\usepackage{cvpr}
\usepackage{times}
\usepackage{epsfig}
\usepackage{graphicx}
\usepackage{amsmath}
\usepackage{amssymb}
\usepackage{color}
\usepackage{xcolor}
\usepackage{algorithm}
\usepackage{algorithmic}
\usepackage{multirow}
\usepackage{colortbl}
\usepackage{comment}
\usepackage{soul}

\definecolor{gd}{RGB}{206,206,206}
\definecolor{gl}{RGB}{240,240,240}

\usepackage[pagebackref=true,breaklinks=true,letterpaper=true,colorlinks,bookmarks=false]{hyperref}
\cvprfinalcopy 

\begin{document}

\title{Self-supervised Modal and View Invariant Feature Learning}

\author{
Longlong Jing$^{1}$
\quad
Yucheng Chen$^{2}$
\quad 
Ling Zhang$^{1}$
\quad 
Mingyi He$^{2}$
\quad
Yingli Tian$^{1}$\thanks{Corresponding author. Email: ytian@ccny.cuny.edu} \\
$^{1}$The City University of New York, ~$^{2}$Northwestern Polytechnical University\\
}

\maketitle

\begin{abstract}

Most of the existing self-supervised feature learning methods for 3D data either learn 3D features from point cloud data or from multi-view images. By exploring the inherent multi-modality attributes of 3D objects, in this paper, we propose to jointly learn modal-invariant and view-invariant features from different modalities including image, point cloud, and mesh with heterogeneous networks for 3D data. In order to learn modal- and view-invariant features, we propose two types of constraints: cross-modal invariance constraint and cross-view invariant constraint. Cross-modal invariance constraint forces the network to maximum the agreement of features from different modalities for same objects, while the cross-view invariance constraint forces the network to maximum agreement of features from different views of images for same objects. The quality of learned features has been tested on different downstream tasks with three modalities of data including point cloud, multi-view images, and mesh. Furthermore, the invariance cross different modalities and views are evaluated with the cross-modal retrieval task. Extensive evaluation results demonstrate that the learned features are robust and have strong generalizability across different tasks.

\end{abstract}

\section{Introduction}

Self-supervised feature learning methods learn visual features from large-scale datasets without requiring any manual annotations. The core of the self-supervised feature learning is to define a pretext task and the visual features are learned through the processing of accomplishing the pretext task. Since it can be easily scaled up to large-datasets, recently some self-supervised methods achieved comparable or even better performance on some downstream tasks than supervised learning methods~\cite{misra2019self, he2019momentum, chen2020simple, asano2019self, jing2019self}.

Most of the existing self-supervised feature learning methods only focus on learning features for one modality. As a rising trend to model 3D visual features, various methods were proposed to learn point cloud features from point cloud either by reconstructing point cloud~\cite{achlioptas2017learning, gadelha2018multiresolution, yang2018foldingnet, zhao20193d}, by generating point cloud with Generative Adversarial Networks~\cite{li2018point, sun2018pointgrow, wu2016learning}, or by accomplishing pretext tasks~\cite{hassani2019unsupervised, zhang2019unsupervised}. Only a few of them~\cite{jing2020self} explored to use cross-modality correspondence of 3D data as supervision signal for 3D self-supervised feature learning.

\begin{figure}[tb]
    \centering
    \includegraphics[width = 0.45\textwidth]{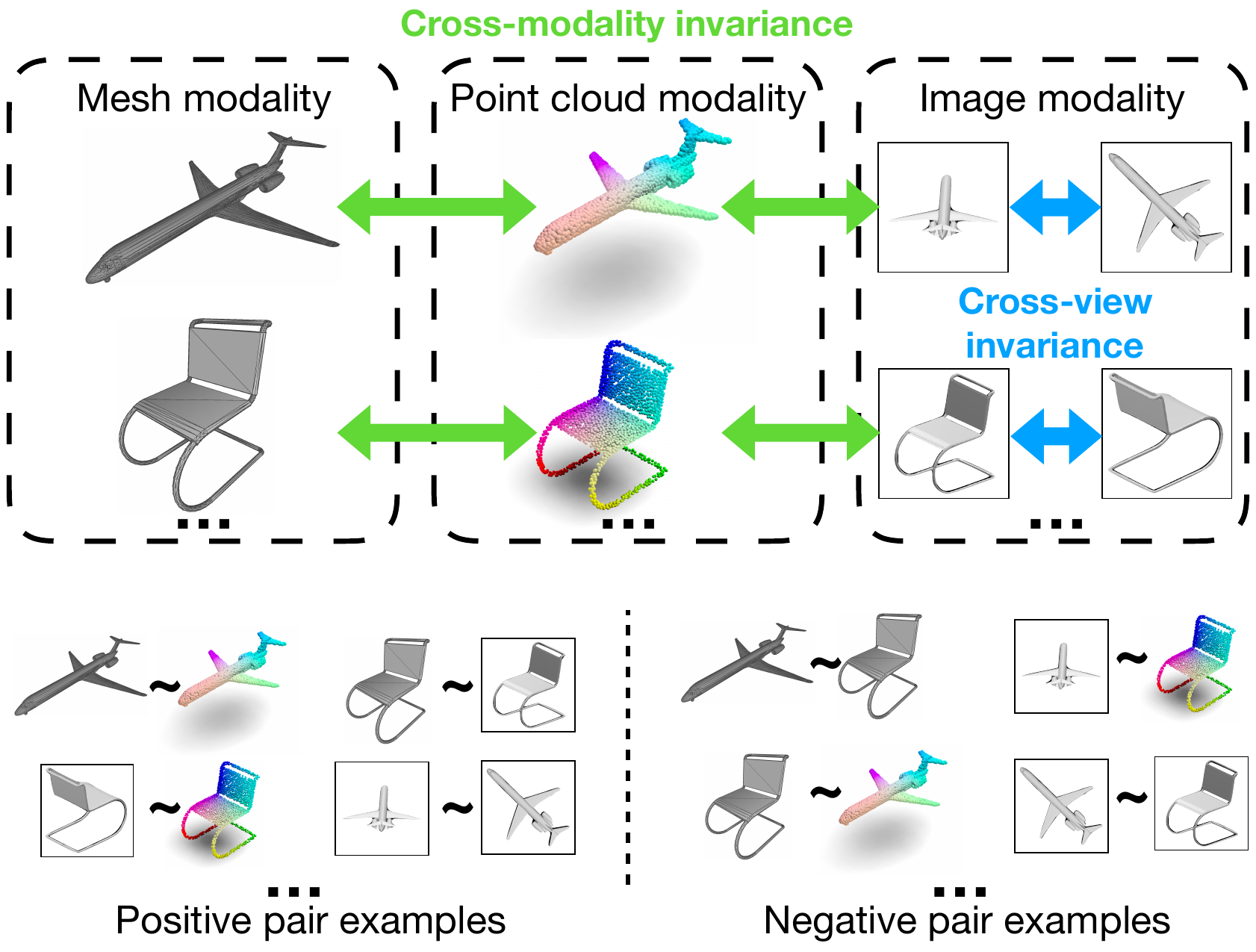}
    \caption{Multi-modality and multi-view representations of 3D objects and pair examples. With cross-modal invariant and cross-view invariant constraints, modal- and view-invariant features can be obtained with heterogeneous networks for different modalities and views of data.}
    \label{fig:motivation}
\end{figure}

Generally 3D data are inherently multi-modalities. Fig.~\ref{fig:motivation} shows different modalities of same objects in mesh, point cloud, and multi-view images. No matter what data format is used to represent an object, its identity remains unchanged. Thus, it is possible to learn features for an object that invariant to its modalities and views. A straightforward idea is to employ the identity invariance of different modalities and views as supervision signals to learn features from unlabeled data. Jing \textit{et al.}~\cite{jing2020self} formulated the identity invariance as a classification task and jointly trained multiple networks to verify whether the inputs from different modalities belong to same object by using the cross-entropy loss. However, the learned features, are not modality-invariant which make it impossible to directly compare the features from different modalities of 3D data.

Recently, contrastive learning has shown great promise and obtained promising performance for recent self-supervised feature learning methods~\cite{chen2020simple, he2019momentum, gordon2020watching}. Similar to triplet loss~\cite{schroff2015facenet}, contrastive loss is to maximize the feature similarity between positive pairs and minimize the feature similarity of other negative pairs. The most commonly used approach to generate positive and negative training pairs is data augmentation on input data~\cite{bachman2019learning, chen2020simple, he2019momentum}. However, dramatic data augmentation, such as cutout and jittering, inevitably changes the distribution of the training inputs far from the real testing data. 

To enable heterogeneous networks to learn features of different modalities and views in same universal space, in this paper, we propose to employ two constraints derived from the attributes of 3D data as supervision signal for self-supervised learning: cross-modal invariance constraint and cross-view modal invariance constraint. The cross-modal invariance constraint forces networks to learn identity features from different modalities of the same object, while the cross-view invariance constraint forces networks to learn an identity features for each object regardless of its view. When jointly learned with the two constraints, modal- and view-invariant features in the universal space can be obtained for each object.

Inspired by the contrastive learning \cite{chen2020simple}, we propose a framework to capture modal-invariant and view-invariant features with heterogeneous networks on three different modalities including mesh, point cloud, and multi-view images. Specifically, the features from different modalities and rendered views of 3D objects are extracted with corresponding subnetworks and combined as positive pairs (sampled with the same object identity) and negative pairs (sampled with different object identities). Then the feature distance of positive cross-modality pairs is minimized, and the feature distance of negative cross-modality pairs is maximized by using contrastive loss. The modal- and view-invariant features are obtained after optimizing networks with the two types of constraints over the positive and negative pairs. The main contributions of this paper are summarized as follows: 

\begin{itemize}
    
    \item We propose a novel self-supervised feature learning schema to jointly learn modal- and view-invariant features for 3D objects end-to-end without using any manual labels.
    
    \item The proposed framework maps the features of different modalities and views of 3D data into same universal space which makes cross-modal retrieval possible for 3D objects.

    \item To the best of our knowledge, we are the first to explore the cross-modal retrieval for 3D objects with multiple modalities in a self-supervised learning way..
    
    \item The effectiveness and generalization of the learned features are demonstrated with extensive experiments with three different modalities on five different tasks  including 3D object recognition, few-shot 3D object recognition, part segmentation, in-domain 3D object retrieval, and cross-modal retrieval.
	
\end{itemize}

\section{Related Work}

\textbf{2D Self-supervised Feature Learning:} Many methods have been proposed to learn visual features from unlabeled 2D data including videos and images. Based on the source of supervision signal, there are four types of self-supervised learning methods: Generation-based method, context-based method, free semantic label-based method, and cross-modal based method. The generation-based methods learn features by reconstructing the data including Auto-encoder, and Generative adversarial networks \cite{GAN}, super resolution \cite{SRGAN}, colorization \cite{colorfulcolorization}, and video future prediction \cite{Self-LSTM}. The context-based methods learn features by using spatial context or temporal context including Jigsaw puzzle \cite{Jigsaw}, geometric transformation \cite{RotNet, jing2018self}, clustering \cite{deepcluster}, frame order reasoning \cite{shuffleandlearn}. The free semantic label-based methods learn features either by data from game engine or to distill features from other unsupervised learning features \cite{watchingmove}. The cross-modal-based methods learn features by the correspondence between a pair of channels of data including video-audio \cite{AVTS} and video-text. Recently, more researchers explore to apply these self-supervised leaning methods to 3D data \cite{jing2020self, zhang2019unsupervised, sauder2019self, hassani2019unsupervised}. 

\textbf{3D Self-supervised Feature Learning:} Several self-supervised learning learning methods have been proposed to learn 3D features for point cloud objects by reconstructing point cloud data ~\cite{achlioptas2017learning,gadelha2018multiresolution,yang2018foldingnet,zhao20193d}, by generating point cloud with GANs ~\cite{li2018point,sun2018pointgrow,thabet2019mortonnet,wu2016learning}, or by training networks to solve pre-defined pretext tasks~\cite{hassani2019unsupervised,sauder2019self,zhang2019unsupervised, jing2020self}. Sauder \textit{et al.} proposed to learn point cloud features by training networks to recognize the relative position of two segments of point cloud \cite{sauder2019self}. Zhang \textit{et al.} designed a  clustering followed by contrastive as pretext task to train networks to learn point cloud features. Hassani \textit{et al.} proposed to train networks with multiple pre-defined pretext tasks including clustering, prediction, and reconstruction for point cloud data \cite{hassani2019unsupervised}. Jing \textit{et al.} proposed to utilize cross-modality relations of point clouds and multi-view images as supervision signal to jointly learn point cloud and image features for 3D objects. However, the point cloud and image features by the network in \cite{jing2020self} are not modality invariant. To thoroughly utilize the cross-modality inherent attributes of 3D data, here we propose to learn modal- and view-invariant features for 3D objects with three different modalities including point cloud, mesh, and images.

\textbf{Contrastive Self-supervised Learning:} The basic principle of contrastive learning, such as Noise Contrastive Estimation (NCE)~\cite{gutmann2010noise}, is to learn representations by contrasting positive and negative pairs. By maximizing the similarity between an anchor sample and a positive sample while minimizing similarity to all other (negative) samples, contrastive learning has shown empirical success in self-supervised learning methods~\cite{bachman2019learning, chen2020simple, he2019momentum, henaff2019data, hjelm2018learning, misra2019self, tian2019contrastive, wu2018unsupervised, gordon2020watching, patrick2020multi} of which the core is to generate positive and negative training pairs by pretext tasks. Most recent work performed training pairs generation in image domain. Studies \cite{bachman2019learning, chen2020simple, he2019momentum} applied dramatic data augmentation such as color jittering, cropping, cutout, and flipping on original images. PIRL~\cite{misra2019self} cropped an image into jigsaw patches, then combines the original image and the shuffled patches into a positive pair. \cite{henaff2019data} divided an image into a grid of overlapping patches and predicts the unseen regions by context patches with Contrastive Predictive Coding (CPC)~\cite{oord2018representation}. \cite{gordon2020watching} extended the contrastive learning on videos by sampling frames from  same video as multiple positive pairs. Contrastive Multiview Coding (CMC)~\cite{tian2019contrastive} used transferred representations (such as Lab color space, depth, and segmentation) of a source image as paired samples. Very few work adopted cross-modality invariance on contrastive learning. \cite{patrick2020multi} generated training pairs by video clips and their corresponding audios. The number of negative pairs affects the probability that positive sample paired with hard negative. Some work~\cite{misra2019self, wu2018unsupervised, gordon2020watching} introduced memory banks storing previous feature outputs to enlarge the negative pairs pool. Momentum Contrast (MoCo)~\cite{he2019momentum} further increased the number of negative samples by a slowly updated negative feature extractor. In this paper, we propose to pair positive and negative object samples by cross-modality invariance in three domains (mesh, point cloud, and image) and cross-view invariance in image domain. 


\section{Method}

\begin{figure*}[tb]
	\centering
	\includegraphics[width = 0.95\textwidth]{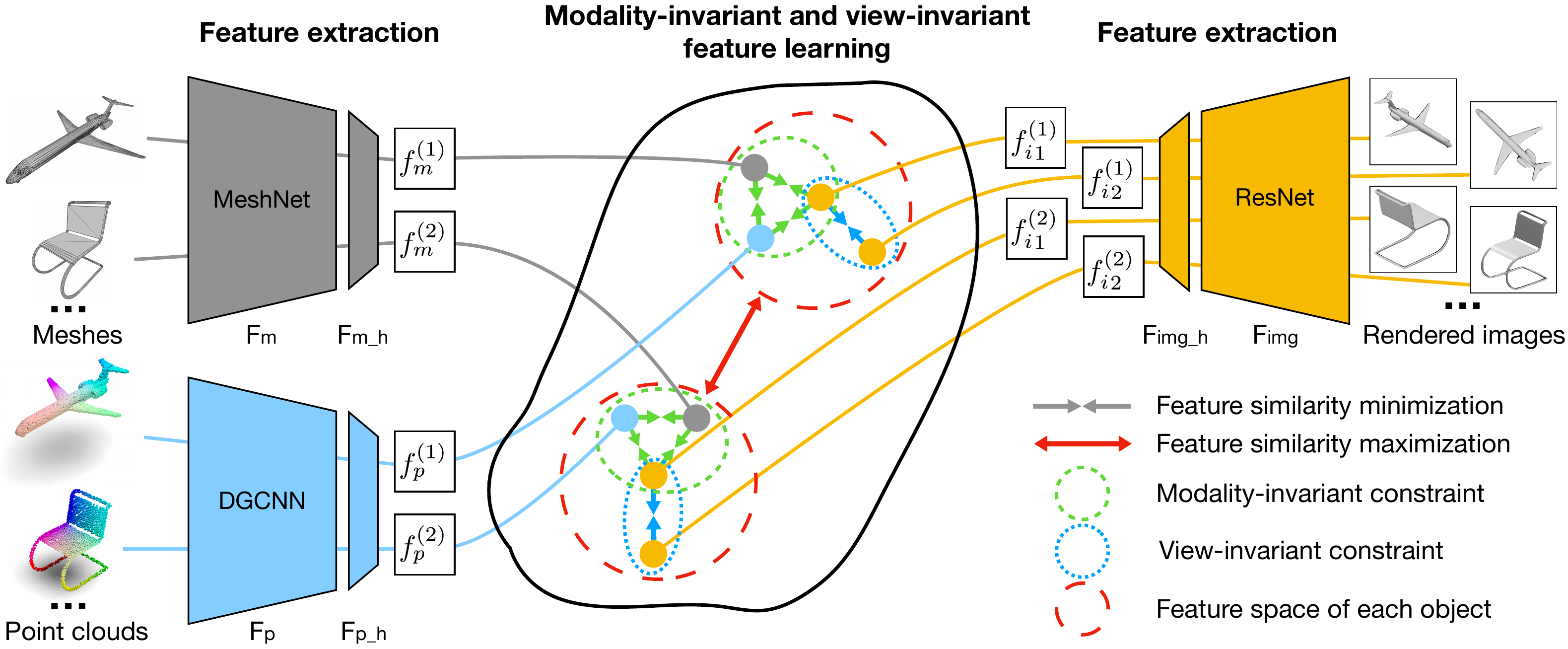}
	\caption{An overview of the proposed self-supervised modal- and view-invariant feature learning framework. Mesh, point cloud, and multi-view image features are extracted by MeshNet, DGCNN, ResNet, and corresponding projection heads, respectively. With contrastive learning to minimum the feature similarity of positive pairs and maximum the feature similarity of negative pairs under modality- and view-invariant constraints, the modal- and view-invariant features can be learned with the proposed heterogeneous framework in the same universal space.}
	\label{fig:framework}
\end{figure*}

An overview of the proposed framework is shown in Fig.~\ref{fig:framework}. The core of our method is to optimize heterogeneous networks to learn modal- and view-invariant features under cross-modality invariance constraint and cross-view invariance constraint by contrasting. Three heterogeneous networks are employed to extract features for three modalities of data including mesh, point cloud, and images, respectively. The framework contains three backbone networks (an image-extracting network $F_{img}$, a point cloud-extracting graph network $F_{p}$, and a mesh-extracting network $F_{m}$) and three corresponding projection heads ($F_{img\_h}$, $F_{p\_h}$, $F_{m\_h}$) mapping features of different modalities into the universal space. Two types of constraints including modality-invariant constraint and view-invariant constraint are used to optimize the network by contrasting paired features of all objects in the universal space. The detailed formulation of our approach is explained in subsection~\ref{sec_param}, and the network architectures are described in subsection~\ref{sec_Architecture}.

\subsection{Model Parameterization}\label{sec_param}

The contrastive learning enables the networks to learn representations by maximizing the similarity between an anchor sample and a positive sample while minimizing similarity to all other (negative) samples, similar to the triplet loss~\cite{schroff2015facenet}. In this paper, the positive pairs are sampled from different modality and view representations with same object identity, while negative pairs with different object identities as shown in Fig~\ref{fig:motivation}. The contrastive learning implementation in this paper uses the following procedure.

Given a batch of $k$ anchor features $\{A^{(1)}, ..., A^{(k)}\}$ and $k$ corresponding positive features $\{P^{(1)}, ..., P^{(k)}\}$. The contrastive loss  for this single anchor-positive batch $l_{ap}$ is defined as
\begin{equation}
\label{eq:defineloss1}
l_{ap} = -\frac{1}{k} \sum_{i=1}^{k} \log\frac{e^{sim(A^{(i)}, P^{(i)})}}{\sum_{j = 1, j \neq i}^{k} e^{sim(A^{(i)}, A^{(j)})} + \sum_{j=1}^{k} e^{sim(A^{(i)}, P^{(j)})}},
\end{equation}
where $sim(A, P)$ denotes the pairwise cosine similarity between two feature vectors as shown in Eq.~\ref{eq:definesim}. $sim(A^{(i)}, P^{(i)})$ calculates the feature similarity of positive pairs, while all other feature similarity calculations are for negative pairs. The optimization of contrastive loss pulls the positive feature pairs closer and pushes the negative feature pairs further in the universe space.

\begin{equation}
\label{eq:definesim}
sim(A, P) = A^\top P/ (\tau \lVert A \rVert \lVert P\rVert),
\end{equation}

 where $\tau$ denotes a temperature parameter. When considering the anchor features as the positive corresponding to original positive features, we can calculate the contrastive loss for the positive-anchor batch $l_{pa}$ as
\begin{equation}
\label{eq:defineloss2}
l_{pa} = -\frac{1}{k} \sum_{i=1}^{k} \log\frac{e^{sim(P^{(i)}, A^{(i)})}}{\sum_{j = 1, j \neq i}^{k} e^{sim(P^{(i)}, P^{(j)})} + \sum_{j=1}^{k} e^{sim(P^{(i)}, A^{(j)})}}.
\end{equation}
Therefore, the complete contrastive loss for the anchor-positive combination is
\begin{equation}
\label{eq:defineloss_final}
L_{AP} = l_{ap} + l_{pa}.
\end{equation}

Let $\mathcal{D} = \{X^{(1)}, ..., X^{(N)}\}$ denotes training data with $N$ specific objects. The $i$-th input sample $X^{(i)} = \{p^{(i)}, m^{(i)}, img_{1}^{(i)}, img_{2}^{(i)}\}$, where  $m^{(i)}$ and $p^{(i)}$ represent the 3D mesh object and the corresponding sampled point cloud, $img_{1}^{(i)}$ and $img_{2}^{(i)}$ are two images under different randomly rendered views generated from the same 3D mesh object. Passing $X^{(i)}$ through the framework, we can obtain corresponding features: $f_p^{(i)}$, $f_m^{(i)}$, $f_{i1}^{(i)}$, and $f_{i2}^{(i)}$. Given a training minibatch $\{X^{(i)}\}_{i=1}^{k}$ of $k$ samples, the positive pairs are derived from same $X^{(i)}$, while the negative pairs are sampled between $X^{(i)}$ and all other samples. In our proposed self-supervised learning schema, two types of feature learning are formulated as supervision signals for contrastive learning to optimize the networks: modal-invariant feature learning and view-invariant feature learning.

\textbf{Modal-invariant Feature Learning:} The object identity information of 3D objects is utilized as the modality-invariant constraint to sample training pairs from different modalities. In this paper, the invariance from three pairs of different data modalities are constructed as training samples including: mesh-point (mesh and point cloud), mesh-image (mesh and image), and point-image (point cloud and image). Taking the mesh and point cloud pair as an example, in a training minibatch of $k$ samples, a batch of $k$ mesh features $\{f_m^{(1)}, ..., f_m^{(k)}\}$ and $k$ corresponding point cloud features $\{f_p^{(1)}, ..., f_p^{(k)}\}$ are considered as the anchor and positive feature batches alternately with each other. Then, following the equations~\ref{eq:defineloss1},~\ref{eq:defineloss2}, and~\ref{eq:defineloss_final}, the complete cross-modal contrastive loss for the mesh and point cloud pair is indicated as $L_{MP}$. The cross-modal contrastive loss for the other two cross-modality pairs (mesh-image and point-image) are calculated in the same way indicated as $L_{MI}$ and $L_{PI}$, respectively.

\textbf{View-invariant Feature Learning: } In image modality, two randomly selected views for each object are used for view-invariant contrastive learning as shown in Fig.~\ref{fig:framework}. With the same loss functions~\ref{eq:defineloss1},~\ref{eq:defineloss2}, and~\ref{eq:defineloss_final}, the cross-view contrastive loss $L_{II}$ is calculated with each image feature with $\{f_{i1}^{(1)}, ..., f_{i1}^{(k)}\}$ and $\{f_{i2}^{(1)}, ..., f_{i2}^{(k)}\}$ in the training minibatch.

When jointly trained with the two constraints, a linear weighted combination of all loss functions is employed to optimize all the networks. The final loss to optimize the framework is as:
\begin{equation}
\label{eq:loss}
\mathcal{L}  = L_{MP} + L_{MI} + L_{PI} + L_{II}.
\end{equation}


The details of the joint training process are illustrated in Algorithm~\ref{alg:training}. After the jointly training finished, three networks $F_{img}$, $F_{p}$, and $F_{m}$ are obtained as pre-trained models for three different modalities. The joint training enables the three feature extractors to map the features from different modalities and views into the same universal space.

\begin{algorithm}[!t]
	\scriptsize
	\caption{The proposed self-supervised feature learning algorithm by contrastive learning cross multimodality and multiviews.}
	\label{alg:training}
	\begin{algorithmic}
		\STATE minibatch size: $k$; 2D image features:$f_{i}$; 3D point cloud features: $f_{p}$; 3D mesh features: $f_{m}$; 
		
		\FOR{\textbf{all} sampled mini-batch $\{X^{(i)}\}_{i=1}^{k}$}
		\STATE \textcolor{gray}{\# feature extraction}
		\STATE \textbf{for all} $i\in \{1, \ldots, k\}$ \textbf{do}
		\STATE $~~~~$ $f_{m}^{(i)} = F_{m\_h}(F_{m}(m^{(i)}))$; $f_{p}^{(i)} = F_{p\_h}(F_{p}(p^{(i)}))$;
		\STATE $~~~~$ $f_{i1}^{(i)} = F_{img\_h}(F_{img}(img_{1}^{(i)}))$; $f_{i2}^{(i)} = F_{img\_h}(F_{img}(img_{2}^{(i)}))$;
		\STATE \textbf{end for}
		
%
%
		
		\STATE \textcolor{gray}{\# loss calculation under modality-invariant constraint}
		\STATE $L_{MP} = l_{mp} + l_{pm}$
		\STATE $L_{MI} = l_{mi} + l_{im}$
		\STATE $L_{PI} = l_{pi} + l_{ip}$
		\STATE \textcolor{gray}{\# loss calculation under view-invariant constraint}
		\STATE $L_{II} = l_{i1i2} + l_{i2i1}$
        \STATE \textcolor{gray}{\# final loss}
		\STATE $\mathcal{L}  = L_{MP} + L_{MI} + L_{PI} + L_{II}$
		\STATE update networks $F_{img}$, $F_{img\_h}$, $F_{p}$, $F_{p\_h}$, $F_{m}$, and $F_{m\_h}$ to minimize $\mathcal{L}$
		\ENDFOR
		\STATE \textbf{return} pre-trained networks $F_{img}$, $F_{p}$, and $F_{m}$
	\end{algorithmic}
\end{algorithm}

\subsection{Framework Architecture}\label{sec_Architecture}

As shown in Fig.~\ref{fig:framework}, the MeshNet~\cite{feng2019meshnet}, dynamic graph convolutional neural network (DGCNN)~\cite{wang2019dynamic}, and ResNet~\cite{he2016deep} are employed as backbone networks to extract representation features from mesh, point cloud, and rendered images, respectively. $F_{m\_h}$, $F_{p\_h}$, $F_{img\_h}$ are three corresponding two-layer fully connected projection heads to map the extracted representation features into an universe space for contrastive learning. The architecture of backbone networks is described as follows.

\textbf{MeshNet:} The backbone architecture for mesh data is MeshNet, denoted as $F_{m}$. MeshNet contains three main blocks: spatial descriptor, structural descriptor, and mesh convolution block. The spatial descriptor applies fully-connected layers (64, 64) to extract spatial features from face’s center. The structural descriptor contains a face rotate convolution within fully-connected layers (32, 32) and (64, 64), and a face kernel correlation with 64 kernels. Two mesh convolution blocks are used to aggregate features with neighboring information which the input/output channels of spatial and structural features are configured as (64, 131, 256, 256) and (256, 256, 512, 512), respectively. After the two mesh convolution blocks, a fully-connected layer (1024) further fuses the neighboring features and a max-pooling layer is employed to extract 512-dimension global features from the aggregated features.

\textbf{DGCNN:} The 3D point cloud feature learning network ($F_{p}$) employs DGCNN as the backbone model due to its capability to model local structures of each point by dynamically constructed graphs. There are four EdgeConv layers and the number of kernels in each layer is $64$, $64$, $64$, and $128$, and the EdgeConv layers aim to construct graphs over $k$ nearest neighbors calculated by KNN and the features for each point are calculated by an MLP over all the $k$ closest points. After the four EdgeConv blocks, a 512-dimension fully connected layer is used to extract per-point features for each point and then a max-pooling layer is employed to extract global features for each object.

\textbf{ResNet:} ResNet18 is employed as the image feature capture network ($F_{img}$) for 2D images. It contains four convolution blocks with a number of \{64, 128, 256, and 512\} kernels. Each convolution block includes two convolution layers followed by a batch-normalization layer and a ReLU layer, except the first convolution block which consists of one convolution layer, one batch-normalization layer, and one max-pooling layer. A global average pooling layer, after the fourth convolution blocks, is used to obtain the global features for each image. Unless specifically pointed out, a 512-dimensional vector after the global average pooling layer is used for all our experiments.

\section{Experimental Results}

\subsection{Experimental Setup}

\textbf{Self-supervised learning:} The proposed framework is jointly trained on ModelNet40 dataset using a SGD optimizer with an initial learning rate of $0.001$, the moment of $0.9$, and weight decay of $0.0005$. The network is optimized with a mini-batch size of $96$ for $160,000$ iterations and the learning rate decrease by $90\%$ every $40,000$ iteration. Data augmentation used for point cloud network includes randomly rotated between [$0$, $2\pi$] degrees along the up-axis, randomly jittered the position of each point by Gaussian noise with zero mean and $0.02$ standard deviation. Data augmentation for images include randomly cropped and randomly flipped with $50$\% probability. Data augmentation for mesh includes random rotation with a degree between [$0$, $2\pi$].

\textbf{Datasets:} Two 3D object benchmarks: ModelNet40~\cite{wu20153d} and ShapeNet~\cite{chang2015shapenet} are used to evaluate the proposed method. The ModelNet40 dataset contains about $12.3k$ meshed models covering $40$ object classes, while about $9.8k$ are used for training and about $2.5k$ for testing. The ShapeNet dataset contains $16$ object categories with about $12.1k$ models for training and about $2.9k$ for testing. 

\textbf{Training data generation:} The point cloud data and multi-view image data are sampled and rendered from same 3D mesh objects, respectively. Specifically, following~\cite{qi2017pointnet}, the point cloud set is sampled from surfaces of mesh objects by Farthest Point Sampling (FPS) algorithm. For each object, uniform $2,048$ points are sampled and normalized into a unit sphere to keep the object shape as much as possible. Same as in~\cite{jing2020self}, we employ Phong reflection model~\cite{phong1975illumination} as the rendering engine to render images from $180$ virtual cameras (viewpoints) to capture perspective of mesh objects  as comprehensive as possible. All virtual cameras are randomly placed along a sphere surface pointing toward the centroid of mesh objects, and one image is rendered form each camera. Note that two of the rendered images are randomly selected for each input training sample.

\textbf{Evaluation of learned 2D and 3D features:} The effectiveness of the self-supervised pre-trained backbone networks $F_{img}$, $F_{p}$, and $F_{m}$ are validated on different downstream tasks for 3D objects including object recognition, few-shot recognition, part segmentation, in-domain and cross-modal retrieval. The features, from the universal space, extracted by the three backbones and corresponding projection heads ($F_{img\_h}$, $F_{p\_h}$, $F_{m\_h}$) are used for the cross-modal retrieval task.

\subsection{Transfer to Object Recognition Tasks}\label{sec_recognition}

The proposed framework can jointly learn features for data with different modalities and views. We validate the effectiveness of the self-supervised pre-trained $F_{img}$, $F_{p}$, and $F_{m}$ on three down-stream supervised tasks: image recognition, point cloud object recognition, and mesh object recognition on the ModelNet40 dataset. Specifically, three linear SVM classifiers are trained based on the extracted features by $F_{img}$, $F_{p}$, and $F_{m}$, respectively. The performance of the SVMS on the testing splits of ModelNet40 dataset are reported and compared. Each feature for the 2D classifier is average from $v$ extracted features of $v$ random views. 

\begin{table}[ht]
	\caption{The performance of object recognition tasks by using self-supervised learned models as feature extractors on the ModelNet40 dataset. "Views" indicates how many views of images are used to obtain the image features.}
	\begin{center}
		\scalebox{1.0}{
			\begin{tabular}{l|c|c|c}
				\hline
				Network     &Modality          &Views   &Accuracy (\%) \\
				\hline
				$F_{img}$   &Image          & 1      & $78.0$\\
				$F_{img}$   &Image          & 2      & $83.1$\\	
				$F_{img}$   &Image          & 4      & $85.8$\\
				$F_{img}$   &Image          & 8      & $87.2$\\
				$F_{img}$   &Image          & 12     & $87.7$\\
				$F_{img}$   &Image          & 36     & $88.2$\\
				\hline
				$F_{p}$     &Point Cloud    & --     & $89.3$\\
				\hline
				$F_{m}$     &Mesh           & --     & $87.7$\\
				\hline
			\end{tabular}
		}
	\end{center}
	\label{tab:recognition}
	\vspace{-15pt}
\end{table}

As shown in Table~\ref{tab:recognition}, the self-supervised pre-trained networks $F_{img}$, $F_{p}$, and $F_{m}$ obtain high accuracy (almost $90$\%) on object recognition tasks with different modalities, showing that the discriminative semantic features are indeed learned through the self-supervised learning process. For the image network, when only one view is available, the performance is only $78$\%, and the performance is significantly boosted when more views are available. Given enough image views, the three networks for the three modalities obtained comparable performance showing that the proposed framework learn robust features for all the modalities. 

\subsection{Transfer to Few-shot Object Recognition Task}\label{sec_fewshot}

To further evaluate the generalization ability of the learned features, we also evaluate the self-supervised pre-trained $F_{img}$, $F_{p}$, and $F_{m}$ on 2D/3D few-shot object recognition tasks and showing the performance in Table~\ref{tab:FewShot} on ModelNet40 dataset. Similar to the settings in Section~\ref{sec_recognition}, three corresponding linear SVM classifiers are trained based on the features of 5, 10, and 20 labeled data for each object category, and the image features for 2D recognition task are generated by max-pooling.

\begin{table}[ht]
	\caption{The performance of few-shot object recognition tasks of the features learned by the proposed self-supervised learning method on ModelNet40 dataset. "S-$\#$" indicates the number of shots for each class.}
	\begin{center}
		\scalebox{1.0}{
			\begin{tabular}{l|c|c|c|c|c}
				\hline
				\multirow{2}*{Network}     &\multirow{2}*{Input}  &\multirow{2}*{Views}  & S-5  & S-10  & S-20\\
				
				&  &    &Acc  &Acc  &Acc\\
				\hline
				$F_{img}$         &Image          & $1$   & $64.0$   & $65.9$   & $70.2$   \\
				$F_{img}$         &Image          & $2$   & $66.0$   & $70.7$   & $75.6$   \\
				$F_{img}$         &Image          & $4$   & $66.0$   & $74.2$   & $77.6$   \\
				$F_{img}$         &Image          & $8$   & $68.4$   & $73.9$   & $78.8$   \\
				$F_{img}$         &Image          & $12$  & $70.4$   & $74.6$   & $79.1$    \\
				\hline
				$F_{p}$         &Point Cloud    & -- & $66.1$  & $71.8$   &$77.8$  \\
				\hline
				$F_{m}$         &Mesh      & -- &$67.9$ &$73.3$ &$79.0$ \\
				\hline
			\end{tabular}
		}
	\end{center}
	\label{tab:FewShot}
	\vspace{-15pt}
\end{table}

As shown in Table~\ref{tab:FewShot}, even only a few labeled data are available for each class, the networks on mesh and point cloud obtain relative high performance for few-shot learning. When only one view of images available, the performance for object recognition with images are much lower than the other two modalities. The performance of object recognition with images is significantly boosted up when more views are available. Overall, the features from mesh modality are more robust than the other two modalities.

\subsection{Transfer to 3D Part Segmentation Task}

For a more thorough effectiveness validation of the learned features across different tasks, we further conduct transfer learning experiments on 3D part segmentation task on the ShapeNet point cloud dataset with a few labeled data available. Since this data only contains labels for the point cloud data, only $F_{p}$ is evaluated on part segmentation task. Four fully connected layers are added on the top of $F_{p}$, and the outputs from all the four EdgeConv blocks and the global features are used to predict the pixel-wise labels. We vary the amount of training data on three experimental setups: (1) with random initialization and supervised training from scratch by the same network~\cite{qi2017pointnet}, (2)  updating parameters on four newly added layers with frozen $F_p$, and (3) fine-tuning parameters  with the pre-trained $F_p$ (unfrozen). The performance is shown in Table~\ref{tab:partseg}. 

\begin{table}[ht]
	\vspace{-8pt}
	\caption{The performance of the three types of settings on different amount of training data from the ShapeNet dataset for object part segmentation task.}
	\begin{center}
		\scalebox{0.99}{
			\begin{tabular}{l|c|c|c|c}
				\hline
				\multirow{2}*{Network}   & Training  & Overall  & Class  & Instance  \\
				&data &  Acc &  mIOU &  mIOU \\
				\hline
				
				Scratch~\cite{qi2017pointnet}  &$1$\%  & $84.4$\%   & $54.1$\%   & $68.0$\% \\
				$F_{p}$-Frozen   &$1$\%  & $86.2$\%  & $58.1$\%  & $70.9$\% \\
				$F_{p}$-Unfrozen &$1$\%  & $88.0$\%   & $60.1$\%   & $73.1$\%\\
				\hline
				Scratch~\cite{qi2017pointnet}  &$2$\%  & $86.6$\%   & $62.7$\%   & $71.9$\% \\
				
				$F_{p}$-Frozen   &$2$\%  & $88.0$\%  & $60.6$\%  & $73.4$\% \\
				$F_{p}$-Unfrozen &$2$\%  & $89.5$\%   & $66.7$\%   & $76.2$\%\\
				\hline
				
				Scratch~\cite{qi2017pointnet}  &$5$\%  & $88.5$\%   & $63.1$\%   & $75.1$\% \\
				
				
				$F_{p}$-Frozen   &$5$\%  & $89.5$\%  & $65.1$\%  & $76.0$\%\\
				$F_{p}$-Unfrozen &$5$\%  & $91.0$\%   & $69.6$\%   & $78.7$\% \\
				\hline
			\end{tabular}
		}
	\end{center}
	\label{tab:partseg}
\end{table}

As shown in Table~\ref{tab:partseg}, for both $F_{p}$-Frozen and $F_{p}$-Unfrozen, the performance of 3D part segmentation can be boosted up in overall accuracy, class mean IOU, and instance mean IOU. Specifically, when only $1$\% labeled data available, the parameter-frozen setup can significantly ($+4\%$ on class mIOU, and $+2.9\%$ on instance mIOU) and increases the performance than training from scratch. It validates that $F_{p}$ is able to learn semantic features from modality-invariant constraints and transfer them across datasets and tasks. As more data are available for training, the overall performance of the network keeps improving and the network significantly benefits from the learned weights. These results suggest that the proposed pretext task lead to learn strong features that are able to be generalized to other tasks.


\subsection{Transfer to Retrieval and Cross-modal Retrieval Tasks}

Compared to all other self-supervised learning models for 3D objects, our method learns modal- and view-invariant features which makes the features of different data modalities be directly comparable. To evaluate the quality of the modal- and view-invariant features, we propose to evaluate them on in-domain and cross-domain retrieval tasks. The performance on the cross-domain retrieval task can show generalizability of the modal-invariance while the performance on the in-domain retrieval task can show generalizability of view-invariance of the learned features. The features for different modalities are extracted by the self-supervised pre-trained backbone networks ($F_{img\_h}$, $F_{p\_h}$, $F_{m\_h}$), and then followed by \textbf{L1} normalization. The Euclidean distance of features is employed to indicate the similarity of two features. All the experiments for in-domain retrieval and cross-domain retrieval tasks are performed on ModelNet40 dataset.

\begin{table}[ht]
	\caption{Performance of  in-domain retrieval tasks with the learned mesh, point cloud, and image features on ModelNet40 dataset. Results of XMV~\cite{jing2020self} are reproduced. The networks with $^*$ are pre-trained on ImageNet dataset.}
	\begin{center}
		\scalebox{0.95}{
			\begin{tabular}{l|c|c|c}
				\hline
				Network  &Source  &Views & mAP\\
				\hline
				$F_m + F_{m\_h}$ & Mesh &--- &62.4\\
				\hline
				XMV~\cite{jing2020self}  & Point Cloud   &---   & 48.4\\
				$F_p + F_{p\_h}$ & Point Cloud  &---  &62.1\\
				
				\hline
				ResNet18~\cite{he2016deep}$^{*}$ & Image  &1  &27.3\\
				XMV \cite{jing2020self} & Image  &1  &30.3\\
				$F_{img} + F_{img\_h}$ & Image &1 &57.9\\
				\hline
				
				ResNet18~\cite{he2016deep}$^{*}$ & Image &2   &34.6\\
				
				XMV \cite{jing2020self} & Image &2 &36.3\\
				
				$F_{img} + F_{img\_h}$   & Image &2  &60.5\\
				\hline
				ResNet18~\cite{he2016deep}$^{*}$  & Image &4  &41.6 \\
				XMV \cite{jing2020self} & Image &4  &46.3\\
				$F_{img} + F_{img\_h}$  & Image &4  &62.3\\
				\hline
			\end{tabular}
		}
	\end{center}
	\label{tab:InDomainRetrieval}
	\vspace{-10pt}
\end{table}

\textbf{In-domain Retrieval:} The performance of three retrieval tasks (image-to-image, point-to-point, and mesh-to-mesh) are shown in Table~\ref{tab:InDomainRetrieval}. For a fair comparison, the networks used in each domain are with the same architecture and trained on the same dataset, except ResNet18~\cite{he2016deep} which is pre-trained on ImageNet dataset. The results from XMV~\cite{jing2020self} are tested by the released pre-trained model. For all the in-domain retrieval tasks, our network achieves relatively high performance and significantly outperforms the recent state-of-the-art self-supervised learning models and the ImageNet pre-trained model. Even when only $1$ image view for each object is available, our model achieves 57.9\% mAP for the in-domain retrieval showing that our model indeed learns view-invariant features. When $4$ views of images for each object is available, the performance of in-domain retrieval task improves by $4.4$\% on ModelNet40 dataset.    

\begin{table}[ht]
	\caption{Performance of \textbf{cross-modal} retrieval tasks with the learned images, point cloud, and mesh features on ModelNet40 dataset.}
	\begin{center}
		\scalebox{0.98}{
			\begin{tabular}{l|c|c|c}
				\hline
				Source &Target &Views &mAP\\
				\hline
				
				Mesh &Point Cloud  &--- &61.6\\
				
				Mesh &Image &1   &58.9\\
				
				Mesh &Image &2   &60.8 \\
				
				Mesh &Image &4   &61.7\\
				
				\hline
				Point Cloud &Mesh   &--- &62.0\\
				Point Cloud &Image  &1   &59.0\\
				
				Point Cloud &Image  &2    &60.8\\
				
				Point Cloud &Image  &4   &61.7\\
				\hline
				
			    Image &Point Cloud  &1   &59.5\\
				
				Image &Point Cloud  &2   &60.7\\

				Image &Point Cloud  &4    &61.7\\
				
				Image &Mesh &1   &59.8\\
				
				Image &Mesh &2   &61.0\\
				
				Image &Mesh &4   &61.7\\
				\hline
			\end{tabular}
		}
	\end{center}
	\label{tab:CrossDomainRetrieval}
\end{table}

\begin{figure*}[tb]
	\centering
	\includegraphics[width =\textwidth]{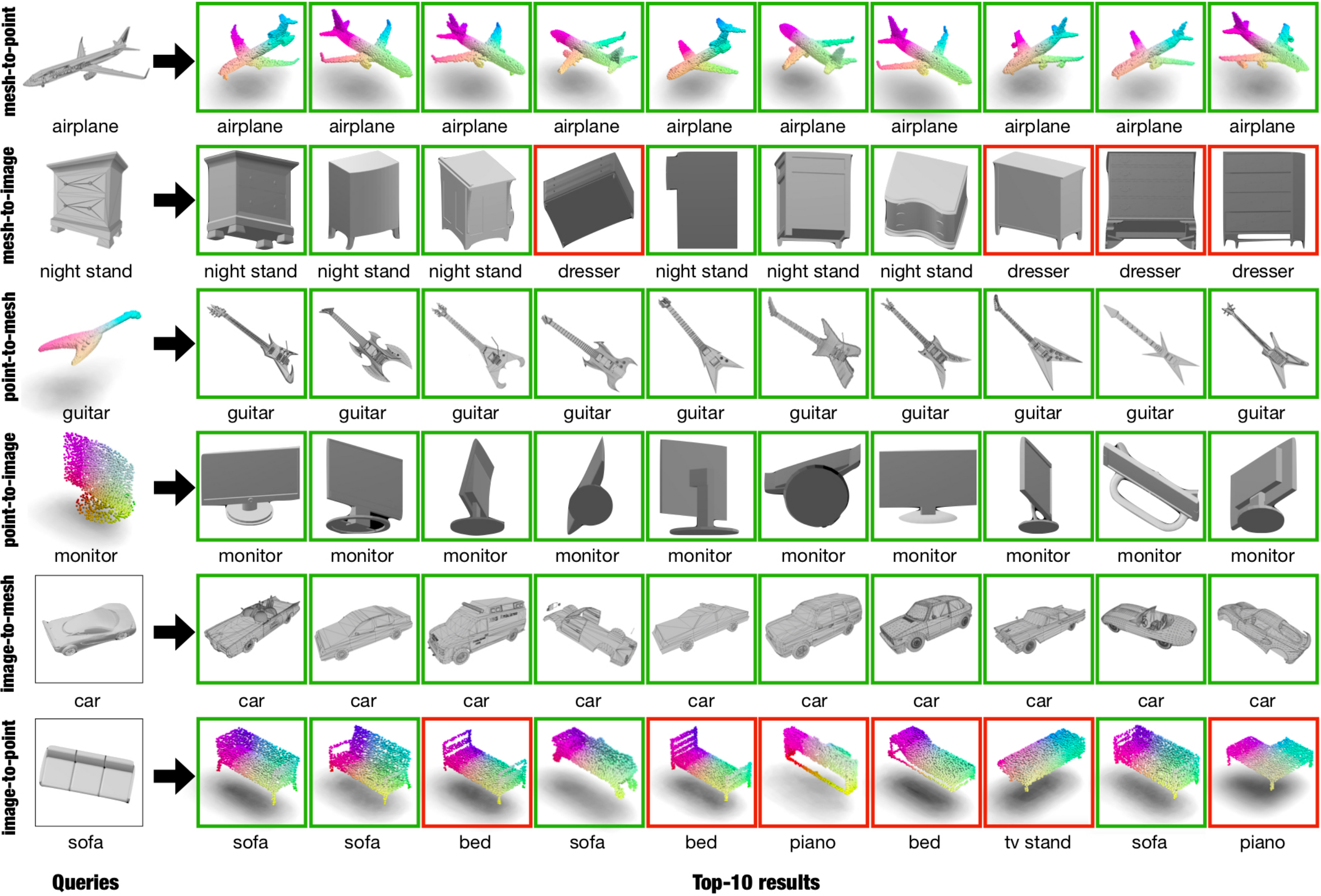}
	\caption{Top-10 ranking lists for six query samples on ModelNet40 dataset by our models. The results with green boundaries belong to the same category as the query, and images with red borders do not.}
	\label{fig:visualization}
\end{figure*}

\textbf{Cross-modal retrieval:} {The jointly learned modal-invariant features in the universal feature space for three different data modalities make the cross-modal retrieval for 3D objects possible, which is, as far as we know, not explored by any other self-supervised or supervised methods.} The cross-modal retrieval task aims to match input data from one modality to different representations from another modality. Here as shown in the Table~\ref{tab:CrossDomainRetrieval}, the feature representation abilities for different modalities are extensively evaluated by six cross-modal retrieval tasks (image-to-point, point-to-image, image-to-mesh, mesh-to-image, mesh-to-point, and point-to-mesh). Our models achieve relatively high performance on all the six different cross-modal combinations showing that the network indeed learns the modal-invariant features. In general, the retrieval accuracy between mesh and point cloud modalities is better than image-involved retrieval due to the inputs of mesh and point cloud contain overall structure information of 3D objects, while a few input views (one or two) are insufficient. When more views (four) are available, the image-involved retrieval accuracy is basically the same. The qualitative visualization results of the top-10 ranking lists for six query samples on ModelNet40 dataset are shown in Fig.~\ref{fig:visualization}. Only one view for each object is used as query or galleries. For objects with unique structures like the airplane, guitar, and car, our models achieves high precision for these classes.

\subsection{Comparison with the State-of-the-art Methods}\label{Comparison}

\begin{table}
	\caption{The comparison with 2D state-of-the-art methods for multi-view 3D object recognition on ModelNet40 dataset.}
	\begin{center}
		\scalebox{1.0}{
			\begin{tabular}{l|c|c|c}
				\hline
				Network  &Supervised &  \#views  &  Acc (\%) \\
				\hline
				MVCNN~\cite{su2015multi} &Yes  & 1 & $85.1$ \\
				DeCAF~\cite{donahue2014decaf}  &Yes 
				& 1 & $83.0$ \\
				Fisher Vector~\cite{sanchez2013image}  &No & 1 & $78.8$ \\
				XMV~\cite{jing2020self} &No & 1 & $72.5$  \\
				$F_{img}$ (Ours) &No  & 1 & $78.0$  \\
				\hline
				DeCAF~\cite{donahue2014decaf} &Yes  & 12 & $88.6$ \\
				MVCNN~\cite{su2015multi} &Yes  & 12 & $88.6$ \\
				XMV~\cite{jing2020self} &No & 12 & $87.3$  \\
				$F_{img}$ (Ours) &No  & 12 & $87.7$  \\
				\hline
			\end{tabular}
		}
	\end{center}
	\label{tab:2D}
\end{table}

\textbf{Object recognition with 2D multi-view images.} The performance of our self-supervised pre-trained $F_{img}$ and the state-of-the-art image-based methods on the ModelNet40 benchmark is shown in Table~\ref{tab:2D}. Methods with multi-view inputs are compared, including hand-crafted model~\cite{sanchez2013image} and supervised feature learning models~\cite{donahue2014decaf, su2015multi, jing2020self}. The setups of our models are same as in subsection~\ref{sec_recognition}. Note that DeCAF~\cite{donahue2014decaf} and MVCNN~\cite{su2015multi} demand large-scale labeled data ImageNet1K for pre-training. XMV~\cite{jing2020self} is pre-trained with two types of modalities (image and point cloud), and the learned features are not modality-invariant. When using same number of views, the performance of our model $F_{img}$ consistently outperforms the state-of-the-art self-supervised learning method \cite{jing2020self} and obtained comparable performance with the supervised methods \cite{donahue2014decaf, su2015multi}. 

\begin{table}[ht]
\caption{The comparison with the state-of-the-art methods for 3D point cloud object recognition on ModelNet40 dataset. * indicates the results in based on mesh modality.}
\begin{center}
	\begin{tabular}{l|c|l|c}
		\hline
		\multicolumn{2}{c|}{Unsupervised} & \multicolumn{2}{c}{Supervised} \\
		\hline
		SPH~\cite{kazhdan2003rotation}  & $68.2$  &PointCNN~\cite{hua2018pointwise}  & $86.1$ \\
		T-L Network~\cite{girdhar2016learning}  & $74.4$
		& PointNet~\cite{li2018point}  & $89.2$ \\
		LFD~\cite{chen2003visual}  & $75.5$   &PointNet++~\cite{qi2017pointnet++}  & $90.7$ \\
		VConv-DAE~\cite{sharma2016vconv}  & $75.5$  & KCNet~\cite{shen2018mining}  & $91.0$   \\
		3D-GAN~\cite{wu2016learning}  & $83.3$  &SpecGCN~\cite{wang2018local} &$91.5$ \\
		Latent-GAN~\cite{achlioptas2017learning}  & $85.7$  & MRTNet~\cite{gadelha2018multiresolution}  & $91.7$  \\
		MRTNet-VAE~\cite{gadelha2018multiresolution}  & $86.4$  & KDNet~\cite{klokov2017escape}  & $91.8$\\
		Contrast~\cite{zhang2019unsupervised} &$86.8$ &MeshNet*~\cite{feng2019meshnet} &$91.9$ \\
		FoldingNet~\cite{yang2018foldingnet}  & $88.4$  & DGCNN~\cite{wang2019dynamic}  & $92.2$ \\
		PointCapsNet~\cite{zhao20193d}  & $88.9$  & &\\
		MultiTask~\cite{hassani2019unsupervised}  & $89.1$  &  &\\
		XMV~\cite{jing2020self} & $89.8$  &   &  \\
		\hline
		$F_{m}$* (Ours)  & $87.7$  &   & \\
		$F_{p}$ (Ours)  & $\textbf{89.3}$  &   &  \\
		\hline
	\end{tabular}
\end{center}
\label{tab:3D}
\end{table}

\textbf{3D object recognition with point cloud and mesh.} Table~\ref{tab:3D} compares the proposed self-supervised pre-trained modelS $F_{p}$ and $F_{m}$ against both self-supervised learning methods~\cite{achlioptas2017learning, chen2003visual, gadelha2018multiresolution, girdhar2016learning, hassani2019unsupervised, jing2020self, kazhdan2003rotation, sharma2016vconv, wu2016learning, yang2018foldingnet, zhao20193d} and supervised learning methods~\cite{feng2019meshnet, gadelha2018multiresolution, hua2018pointwise, klokov2017escape, li2018point, qi2017pointnet++, shen2018mining,wang2018local, wang2019dynamic} on the ModelNet40 benchmark. Our self-supervised learning approach achieves comparable performance to the supervised methods on the ModelNet40 dataset. The performance of our model are very close ($0.5\%$ lower) to previous self-supervised learning methods, while the learned modal-invariant features are applicable for more downstream tasks such as retrieval task. Worth to note that most of the self-supervised learning only learn features for point cloud data, while our method is able to learn modal- and view-invariant features for different modalities, and it can be easily extend to other modalities. 

\section{Conclusion}

In this paper, we have proposed a novel self-supervised learning method to jointly learn features which are invariant to different  modalities and views. Different from all the previous self-supervised learning methods, our method is able to learn features for different modalities in the same universal space which makes it possible to explore a new task, i.e., cross-modal 3D object retrieval. The image features, mesh features, and point cloud features learned by three different networks have been extensively tested on different tasks including 3D object recognition, few-shot learning, part segmentation, 3D objects retrieval and cross-modal retrieval, and demonstrated the strong generalizability of the learned features.

\section{Acknowledgement}
This material is partially based upon the work supported by National Science Foundation (NSF) under award number IIS-1400802.

{\small
\bibliographystyle{ieee_fullname}
\bibliography{Contrastive}
}
\end{document}